\documentclass[10pt,letterpaper]{article}

\usepackage{cogsci}

\usepackage{booktabs}       %
\usepackage{amsfonts}       %
\usepackage{nicefrac}       %
\usepackage{microtype}      %
\usepackage{bbm}

\cogscifinalcopy %

\usepackage{pslatex}
\usepackage{apacite}
\usepackage{float} %

\usepackage{graphicx}
\usepackage{subcaption}

\usepackage[dvipsnames]{xcolor}

\title{A Computational Model of Early Word Learning from the Infant's Point of View}
 
\author{
\\{\large \bf Satoshi Tsutsui$^{1}$, Arjun Chandrasekaran$^{3}$, Md Alimoor Reza$^{1}$, David Crandall$^{1}$, Chen Yu$^{2}$}\\
\{stsutsui,mdreza,djcran,chenyu\}@iu.edu, arjun.chandrasekaran@tuebingen.mpg.de
\\
$^{1}$ Luddy School of Informatics, Computing, and Engineering, Indiana University, USA \\
$^{2}$ Department of Psychological and Brain Sciences, Indiana University, USA \\
$^{3}$ Max Planck Institute for Intelligent Systems, Germany 
\\}

\begin{document}

\maketitle

\begin{abstract}
Human infants have the remarkable ability to learn the associations
between object names and visual objects from inherently ambiguous
experiences. Researchers in cognitive science and developmental
psychology have built formal models that implement in-principle
learning algorithms, and then used pre-selected and pre-cleaned
datasets to test the abilities of the models to find statistical
regularities in the input data. In contrast to  previous modeling
approaches, the present study used egocentric video and gaze data
collected from infant learners during natural toy play with their parents.
This allowed us to capture the learning
environment from the perspective of the learner's own point of
view. We then used a Convolutional Neural Network (CNN) model to
process sensory data from the infant's point of view and learn
name-object associations from scratch.  As the first model that takes raw
egocentric video to simulate infant word learning, the present study
provides a proof of principle that the problem of early word learning
can be solved, using actual visual data perceived by infant
learners. Moreover, we conducted simulation experiments to
systematically determine how visual, perceptual, and attentional
properties of infants' sensory experiences may affect word
learning.

\end{abstract}

\section{Introduction}

Infants show knowledge of their first words as early as 6 months old
and produce their first words at around a year. 
Learning object names --- a major component of their early vocabularies --- in everyday contexts requires young
learners to not only find and recognize visual objects in view but also to map
them with heard names. In such a context, infants seem to be able to learn from a
sea of data relevant to object names and their referents because
parents interact with and talk to their infants in various occasions
--- from toy play, to picture book reading, to family meal time~\cite{yu2012embodied}.

\begin{figure}[tb!]
\centering
	\includegraphics[width=0.75\columnwidth]{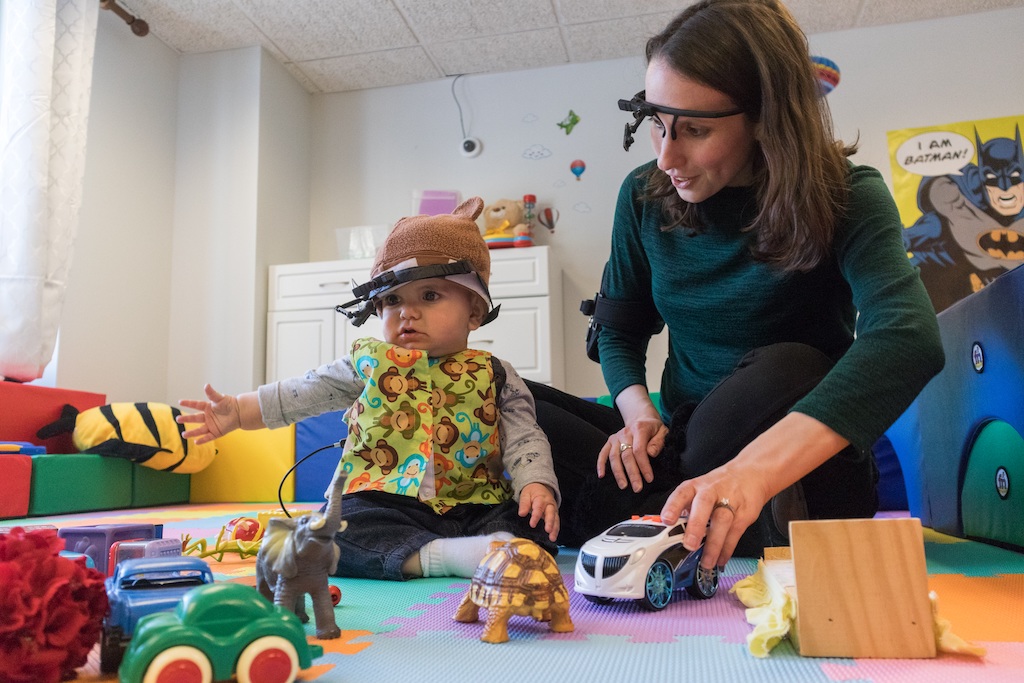}
	\caption{An infant and parent play with a set of toys in a
          free-flowing joint play session. Both participants wore 
          head-mounted cameras and eye trackers to record egocentric video and gaze
          data from their own perspectives. \label{fig:env}}
\end{figure}

\begin{figure*}[htb!]
	\centering
    \includegraphics[width=\textwidth]{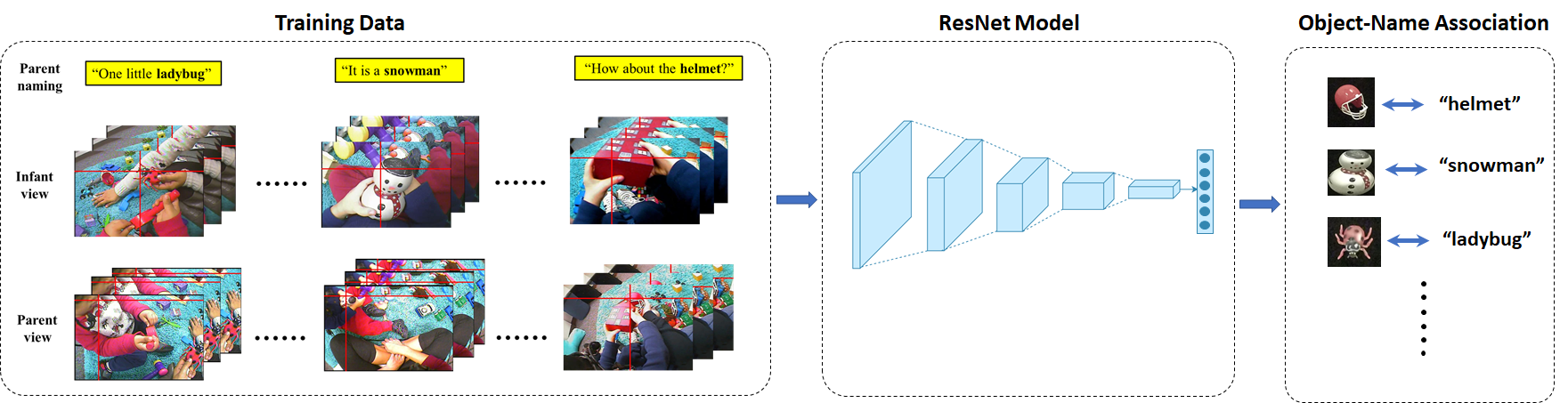}
    \caption {Overview of our approach. The training data were created by extracting egocentric image frames around the moments when parents named objects in free-flowing interaction. The data was fed into deep learning (ResNet) models to find and associate visual objects in view with names in parent speech. As a result, the models built the associations between heard labels and visual presentations of target objects. 
    \label{fig:overview} }
\end{figure*}

However, if we take the young learner's point of view, we see that the
task of word learning is quite challenging. Imagine an infant and
parent playing with several toys jumbled together as shown in
Figure~\ref{fig:env}. When the parent names a particular toy at a
particular moment, the infant perceives 2-dimensional images on the
retina from a first-person point of view, as shown in
Figure~\ref{fig:overview}. These images usually contain multiple
objects in view. Since the learner does not yet know the name of the
toy, how do they recognize all the toys in view and then infer the
target to which the parent is referring? This \textit{referential
  uncertainty}~\cite{quine1960word} is the classic puzzle of early
word learning:
because real-life learning situations are replete with objects
and events, a challenge for young word learners is to recognize and
identify the correct referent from many possible candidates at a given
naming moment. Despite many experimental studies on 
infants~\cite{golinkoff2000becoming} and much computational work on simulating early
word learning~\cite{yu2007unified, frank2009}, how young children solve this problem 
remains an open question.

Decades of research in developmental psychology and cognitive science
have attempted to resolve this mystery.  Researchers have designed
human laboratory experiments by creating experimental training
datasets and testing the abilities of human learners to learn from
them~\cite{golinkoff2000becoming}. In computational studies,
researchers have built models that implement in-principle learning
algorithms, and created training sets to test the abilities of the
models to find statistical regularities in the input data. Some
work in modeling word learning has used sensory data collected
from adult learners or robots~\cite{Roy2002,yu2007unified,Rasanen2019}, while many models take symbolic data
or simplified
inputs~\cite{frank2009,kachergis2017,smith2011,fazly2010,yu2007unified}.
Little is known about whether these models can scale up to address the
same problems faced by infants in real-world learning. As recently
pointed out in~\cite{dupoux2018}, the research field of cognitive
modeling needs to move toward using realistic data as input because
all the learning processes in human cognitive systems are sensitive to
the input signals~\cite{smith2018}. If our ultimate goal is to
understand how infants learn language in the real world --- not in
laboratories or in simulated environment --- we should model internal
learning processes with natural statistics of the learning
environment. This paper takes a step towards this goal and uses
data collected by infants as they naturally play with toys and interact with parents.

Recent advances in computational and sensing techniques (deep
learning, wearable sensors, etc.)  could
revolutionize the study of cognitive modeling. In the field of machine
learning, Convolutional Neural Networks (CNNs) have achieved
impressive learning results and even outperform humans on some specific
tasks~\cite{silver2016mastering,he2015delving}. In the field of computer vision,
small wearable cameras have been used to capture an
approximation of the visual field of their human wearer. Video from
this egocentric point of view provides a unique perspective of the
visual world that is inherently human-centric, giving a level of
detail and ubiquity that may well exceed what is possible from
environmental cameras in a third-person point-of-view. Recently,
head-mounted cameras and eye trackers have been used in developmental
psychology to collect fine-grained information about what infants are
seeing and doing in real
time~\cite{he2015delving,silver2016mastering}. These new
technologies make it feasible to build
computational models using inputs that are very close to infants' actual sensory
experiences, in order to understand the rich complexity of
infants' sensory experiences available for word learning.

In the present study, we collect egocentric video and gaze data from
infant learners as they and their parents naturally play with a set of
toys. This allows us to capture the learning environment from the
perspective of the learner's own point of view. We then build a
computational system that processes this infant sensory data 
to learn name-object associations from scratch. As the
first model taking raw egocentric video to simulate infant word
learning, the present study has two primary goals. The first aim is to
provide a proof of principle that the problem of early word learning
can be solved using raw data. The second aim is to systematically
determine the computational roles of visual, perceptual, and
attentional properties that may influence word learning. This
examination allows us to generate quantitative predictions which can
be further tested in future experimental studies.

\section{Method}
\subsection{Data Collection}
To closely approximate the input perceived by infants, we collected
visual and audio data from everyday toy play --- a context in which
infants naturally learn about objects and their names.  We developed
and used an experimental setup in which we placed a camera on the
infant's head to collect egocentric video of their field of view,
as shown in
Figure~\ref{fig:env}.  We also used a head-mounted eye gaze tracker to
record their visual attention. 
Additionally, we collected synchronized video and gaze data from the
parent during the same play session.

Thirty-four child-parent dyads participated in our study.  Each dyad
was brought into a room with 24 toys (the same as
in~\cite{bambach2018toddler}) scattered on the floor.  Children and
parents were told to play with the toys, without more specific
directions.  The children ranged in age from 15.2 to 24.2
months ($\mu$=19.4 months, $\sigma$=2.2 months). We collected five
synchronized videos per dyad (head camera and eye camera for child,
head camera and eye camera for parent, and a third-person view camera
-- see Figure~\ref{fig:env}). The final dataset contains 212
minutes of synchronized video, with each dyad contributing different amounts of
data ranging from 3.4 minutes to 11.6 minutes ($\mu$=7.5 minutes,
$\sigma$=2.3 minutes).  The head-mounted eye trackers recorded video
at 30 frames per second and 480 $\times$ 640 pixels per frame, with a
horizontal field of view of about 70 degrees. We followed validated
best practices for mounting the head cameras so as to best approximate
participants' actual first-person views, and for calibrating the eye
trackers~\cite{slone2018gaze}.
\subsection{Training Data} 
Parents' speech during toy play was fully transcribed and divided into
spoken utterances, each defined as a string of speech between two
periods of silence lasting at least
400ms~\cite{yu2012embodied}. Spoken utterances containing the name of
one of the objects were marked as ``naming utterances'' (e.g. ``that's
a helmet''). For each naming utterance, trained coders annotated the
intended referent object. On average, parents produced 15.51
utterances per minute ($\sigma$=4.56), 4.82 of which were referential
($\sigma$=2.09). In total, the entire training dataset contains 1,459
naming utterances.

Recent studies on infant word learning show that the moments during 
and after hearing a word are critical for young learners to associate 
seen objects with heard words~\cite{yu2012embodied}. In light of this, 
we temporally aligned speech data with video data, and used a 3-sec 
temporal window starting from the  onset of each naming utterance. 
Given that each naming utterance lasted about 1.5 to 2 seconds, a 3-sec 
window captured both the moments that infants heard  the target name in 
parent speech and also the moments after hearing the name. For each 
temporal window, a total of 90 image frames (30 frames per second) were extracted.
To summarize, the final training dataset consists of all the naming instances 
in parent-child joint play, with each instance  containing a target name 
and a set of 90 image frames from the child's first-person camera that 
co-occur with the naming utterance. As shown in Figure~\ref{fig:overview}, 
each image typically contains multiple visual objects 
and the named object may or may not be in view. 

\begin{figure}[htb!]
	\includegraphics[width=\linewidth]{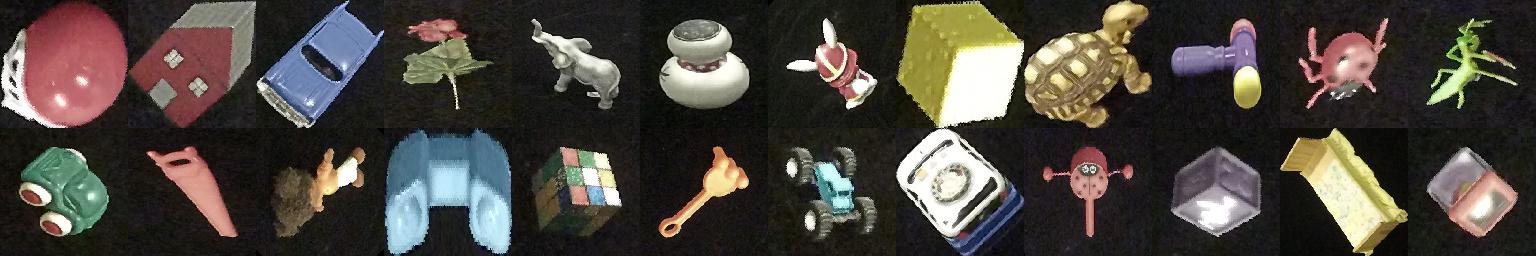}%
	\caption{Testing images. We evaluate the models trained from egocentric images using systematically captured images from various views with a clean background.\label{fig:test}}
\end{figure}

\subsection{Testing Data and Evaluation Metrics} 
To evaluate the result of word learning, we prepared a separate set of
clean canonical images for each of the 24 objects varying in camera
view and object size and orientation in a similar manner to previous
work~\cite{bambach2016active}. In particular, we took pictures of each
toy from eight different points of view (45 degree rotations around
the vertical axis), totaling 3,072 images (see
Fig~\ref{fig:test}). This test set allowed us to examine whether the
models generalized the learned names to new visual instances never
seen before.  During test, we presented one image at a time to a
trained model and checked whether the model generated the correct
label. We compute mean accuracy (i.e., the number of correctly
classified images over the total number of test images) as the
evaluation metric.

\subsection{Simulating acuity}
\begin{figure}[t]
	\centering
	\begin{subfigure}[b]{0.49\linewidth}
		\includegraphics[width=\linewidth]{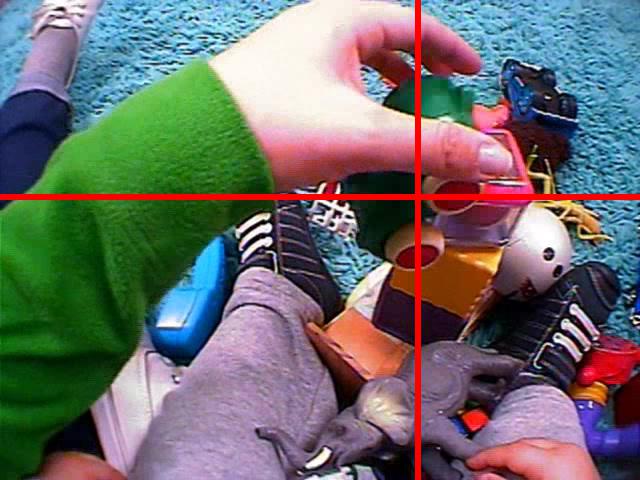}
		\caption{Original\label{fig:original}}
	\end{subfigure}
    \hfill
	\begin{subfigure}[b]{0.49\linewidth}
		\includegraphics[width=\linewidth]{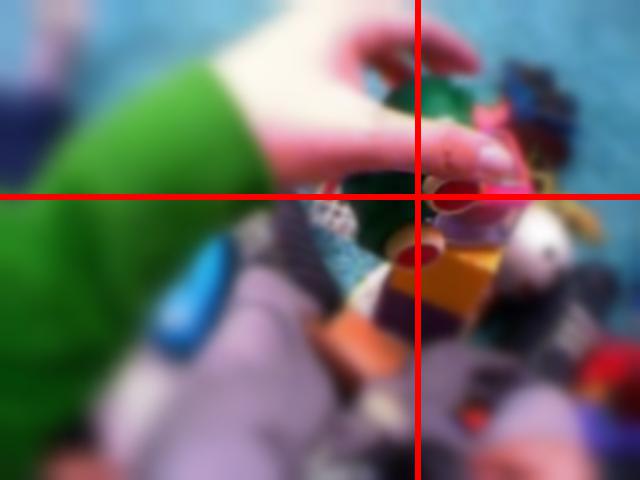}
		\caption{Acuity simulation\label{fig:acuity}}
	\end{subfigure}
	\caption{ We simulated foveated vision by applying an acuity filter to the original egocentric image, based on the eye gaze position (red crosshairs).  \label{fig:basedata}}
\end{figure}

Egocentric video captured by head-mounted cameras provides a good
approximation of the field of view of the infant. However, the human
visual system exhibits well-defined contrast sensitivity due to retinal
eccentricity: the area centered around the gaze point (the fovea)
captures a high-resolution image, while the imagery in the periphery
is captured at dramatically lower resolution due to its lesser
sensitivity to higher spatial frequencies. As a result, the human
visual system does not process all ``pixels'' in the first-person
image equally, but instead focuses more on the pixels around the
fovea. To closely approximate the visual signals that are ``input'' to
a learner's learning system, we implemented the method of~\cite{perry}
to simulate the effect of foveated visual acuity on each frame.  The
basic idea is to preserve the original high-resolution image at the
center of gaze while increasing blur progressively towards the periphery,
as shown in Figure~\ref{fig:acuity}. This technique applies a model of
what is known about human visual acuity and has been validated with
human psychophysical studies~\cite{perry}.

\subsection{Convolutional Neural Networks Models}
We used a state-of-the-art CNN model, ResNet50~\cite{he2016resnet},
trained with stochastic gradient descent (SGD). The network outputs a
softmax probability distribution over 24 object labels, so the 
label with the highest probability is the predicted object. SGD 
optimizes the CNN parameters to minimize the cross entropy loss 
between the predicted distribution and the ground truth (one-hot) distribution. 
Before SGD, we initialized the parameters of ResNet50 with a model
pretrained on ImageNet~\cite{russakovsky2015imagenet}. Thus, the model
can reuse the visual filters learned on ImageNet to avoid having to
learn the low-level visual filters from scratch.  The training
images were resized to $224 \times 224$ pixels with bilinear interpolation.  We
used SGD with batch size 128, momentum $0.9$,
and initial learning rate 0.01. We decreased the learning rate by a
factor of 10 when the performance stopped improving, and ended
training when the learning rate reached 0.0001. Because training was
stochastic, there is  natural variation across training runs; we
thus ran each of our experiments 10 times and report means and
standard deviations.  Moreover, since our goal was to discover general
principles that lead to successful word learning and not to analyze
the results of individual objects, we applied a mixed-effect
logistic regression with random effects of trial and object in each of
our analyses.

\section{Experiments and Results}

\subsection{Study 1: Learning object names from raw egocentric video}

The aim of Study 1 is to demonstrate that a state-of-the-art machine
learning model can be trained to associate object names with visual
objects by using egocentric data closely approximating sensory
experiences of infant learners.  We also evaluated models learned with
parent view data in order to compare the informativeness of these
different views.  Moreover, to examine the impact of properties of the
training data, we created several simulation conditions by
sub-sampling the whole set of 1459 into seven subsets with different numbers
of naming events (50, 100, 200, 400, 600, 800, 1100). While we expected
that more naming instances would lead to better learning, 
we sought to quantify this relationship.

\begin{figure}[t]
\begin{center}
	\includegraphics[width=0.9\linewidth]{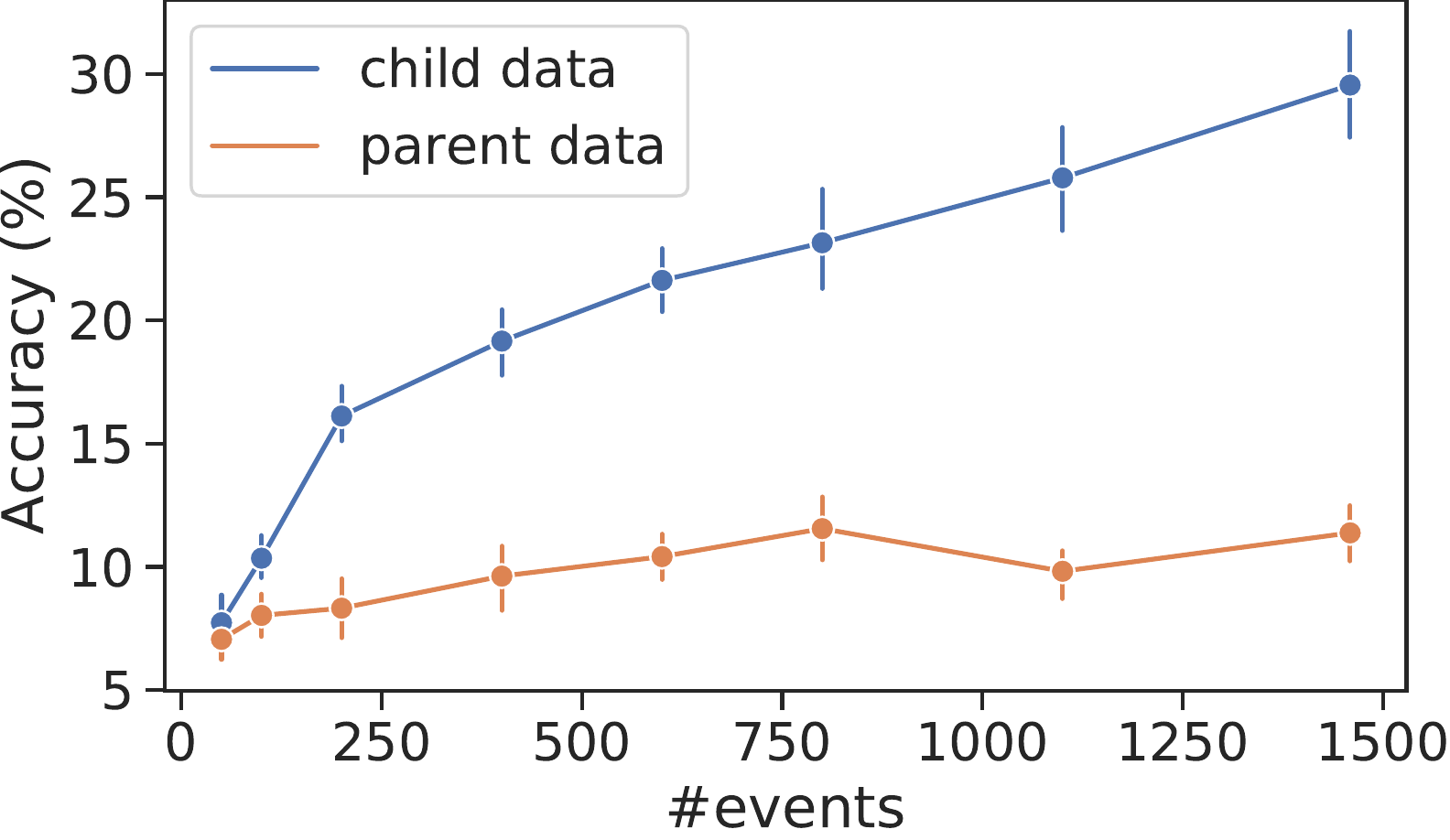}
\end{center}
	\caption{Results from models trained with infant data
          improve with more naming instances, while the
models trained with the parent data show
          no improvement. \label{fig:study1}}
\end{figure}

 Figure \ref{fig:study1} reveals two noticeable patterns in the
 models trained on the infant data and the model trained on the parent
 data. First, when there are 200 or more naming events, 
 models trained with infant data consistently outperformed
 the same models trained on parent data (e.g., for 200 naming
 events: $M_{infant}=16.12\%, SE_{infant}=1.73\%; M_{parent}=8.32\%,
 SE_{parent}=1.27\%; \beta=0.34, t=3.64,p<0.001$).  Second, as the
 quantity of training data increased, the models trained on infant
 data obtained better performance while the models trained on the
 parent data saturated. Taken together, these results 
 provide convincing evidence that the model can solve the name-object
 mapping problem from raw video, and that the infant data contain
 certain properties leading to better word learning. The finding that
 infant data lead to better learning is consistent with recent results
 reported on another topic in early development -- visual object
 recognition~\cite{bambach2018toddler}.

\subsection{Study 2: Examining the effects of different attentional strategies} 
Humans perform an average of approximately three eye movements per second because
our visual systems actively select visual information 
which is then fed into internal cognitive and learning
processes. Thus during the 3-second window during and after hearing a naming
utterance, an infant learner may generate multiple looks on different objects
in view, or, alternatively, 
may sustain their attention on one object.
The aim of Study 2 is to investigate whether 
different attention strategies during naming events influence
word learning, and if so, in which ways.

\begin{figure}[t]
        \centering
	    \includegraphics[width=\linewidth]{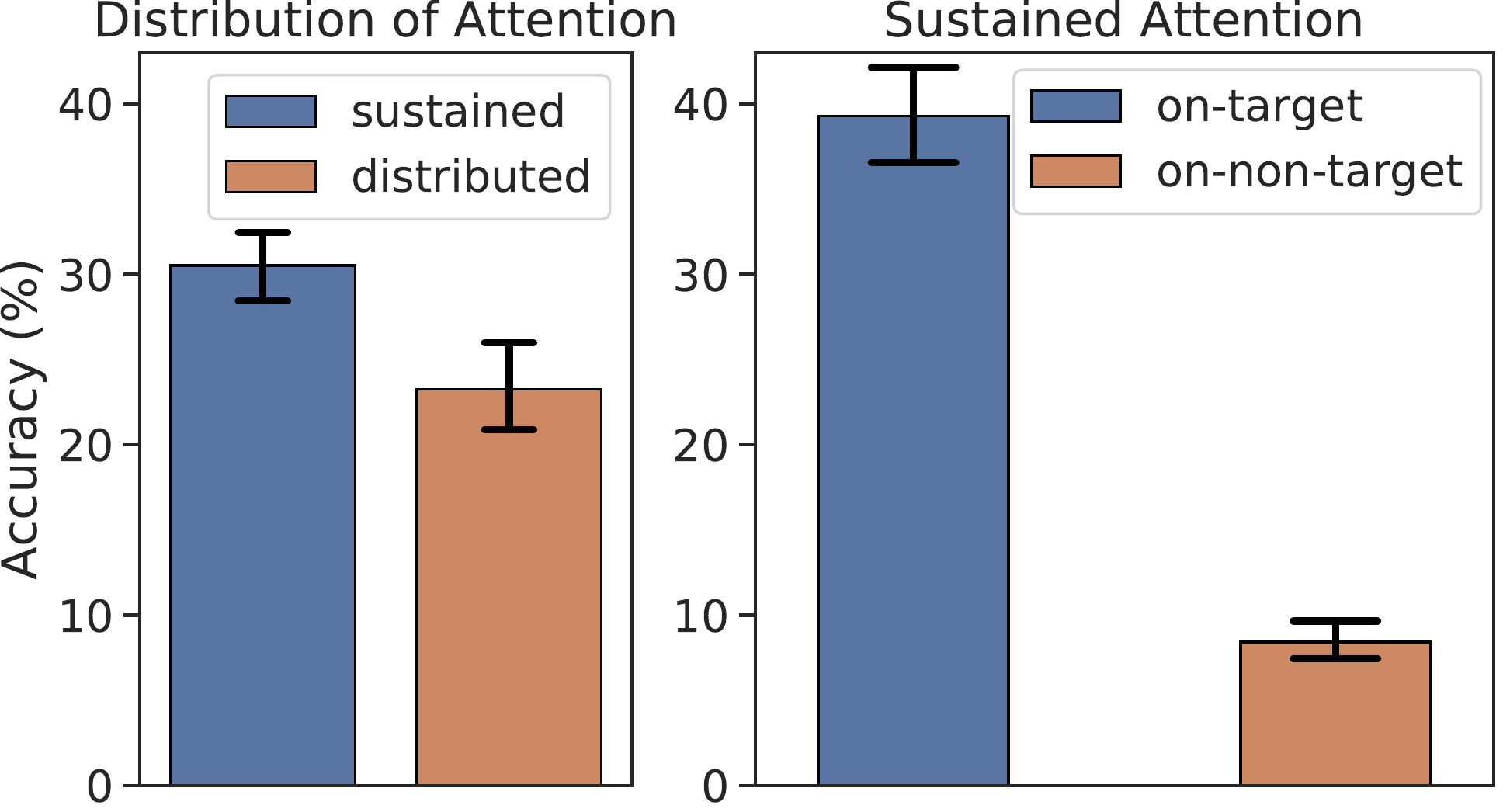}
	\caption{The model trained with sustained attention events
          outperformed the model trained with distributed attention
          events. Within the sustained attention events, the model
          trained with on-target instances outperformed the model
          trained with on-non-target
          instances. \label{fig:study3temporalAtt}}
\end{figure}

To answer this question, we first assigned each naming
event into one of two categories:
\textit{sustained attention} if the infant
attended to a single object for more than 60\% of the
frames in the naming event, and \textit{distributed attention} otherwise.
This split resulted in 750 sustained attention (SA) and 709
distributed attention (DA) events. In either case, the infant may or may
not attend to the named object because the definition is based on the
distribution of infant attention, \textit{not} on which objects were
attended in a naming event. We trained two identical models, one on SA
instances and one on DA instances. The results in Figure~\ref{fig:study3temporalAtt} reveal that the model trained with 
sustained attention events ($M_{sustained}=30.53\%,
SE_{sustained}=2.08\%$) outperformed the  model trained with 
distributed attention events ($M_{distributed}=23.26\%,
SE_{distributed}=1.78\%; \beta=0.20, t=2.65,p<0.005$), suggesting that
sustained attention on a single object while hearing a name leads to
better learning.

Of course, infants may or may not show sustained attention on the
object actually named in parent speech.
In total,
infants attended to the target in 452 out of 750 SA events, and
attended to a non-target object in the other 298 SA events. Attending
to the target object with sustained attention should help learning
while sustained attention on a non-target object should hinder
learning. To test this prediction,
we sub-sampled 298 on-target events from 452 SA events, and compared
them with the remaining 298 on-non-target events. As shown in
Figure~\ref{fig:study3temporalAtt}, the model trained with the on-target
events ($M_{target}=39.27\%, SE_{target}=2.30\%$) achieved
significantly higher accuracy than the model trained on 
on-non-target events ($M_{non-target}=8.42\%, SE_{non-target}=1.20\%;
\beta=0.98, t=14.52,p<0.001$).

In everyday learning contexts such as toy play, young learners do not
passively perceive information from the environment; instead, the
visual input to internal learning processes is highly selective
moment-to-moment. The ability to sustain attention in such contexts
is critical for early development and has been linked to healthy
developmental outcomes~\cite{ruff2001attention}. The results from the
present study suggest a pathway through which sustained attention
during parent naming moments creates sensory experiences that
facilitate word learning.
 
\subsection{Study 3: Examining the effects of visual properties of attended objects}
One effect of sustained attention during a naming moment is to
consistently select a certain area in the egocentric view so that the
learning system can process the visual information in that
focused area to find the target object and link it with the heard
label. Moving from the attentional level to the sensory level, we
argue that associating object names with visual objects starts with
visual information selected in the infant's egocentric view, and
therefore the factors that matter to word learning may not just be
attended objects  but sensory information selected and
processed in the naming moments. 
Study 3 seeks to determine how visual
properties of attended objects influence word learning.

Previous studies using head-mounted cameras and head-mounted eye
trackers showed that visual objects attended by infants tend to
possess certain visual properties --- e.g., they tend to
be large in view, which provides a high-resolution image of the object~\cite{yu2012embodied}. In light of this, the present simulation
focused on object size.
The naming events were grouped into two subsets by a median split of
object size. The large subset contains naming instances in which named
objects are larger than the median size (6\%) whereas the small subset
contains naming instances in which named objects are smaller than the
median. The same model was separately trained on the large set and the
small set. We found the model trained with large objects achieved
significantly higher accuracy on the test dataset than that trained
with small objects ($M_{large}=30.50\%, SE_{large}=2.20\%;
M_{small}=18.81\%, SE_{small}=1.81\%; \beta=0.29, t=4.12,p<0.001$).

\begin{figure}[t]
    \centering
		\includegraphics[width=\linewidth]{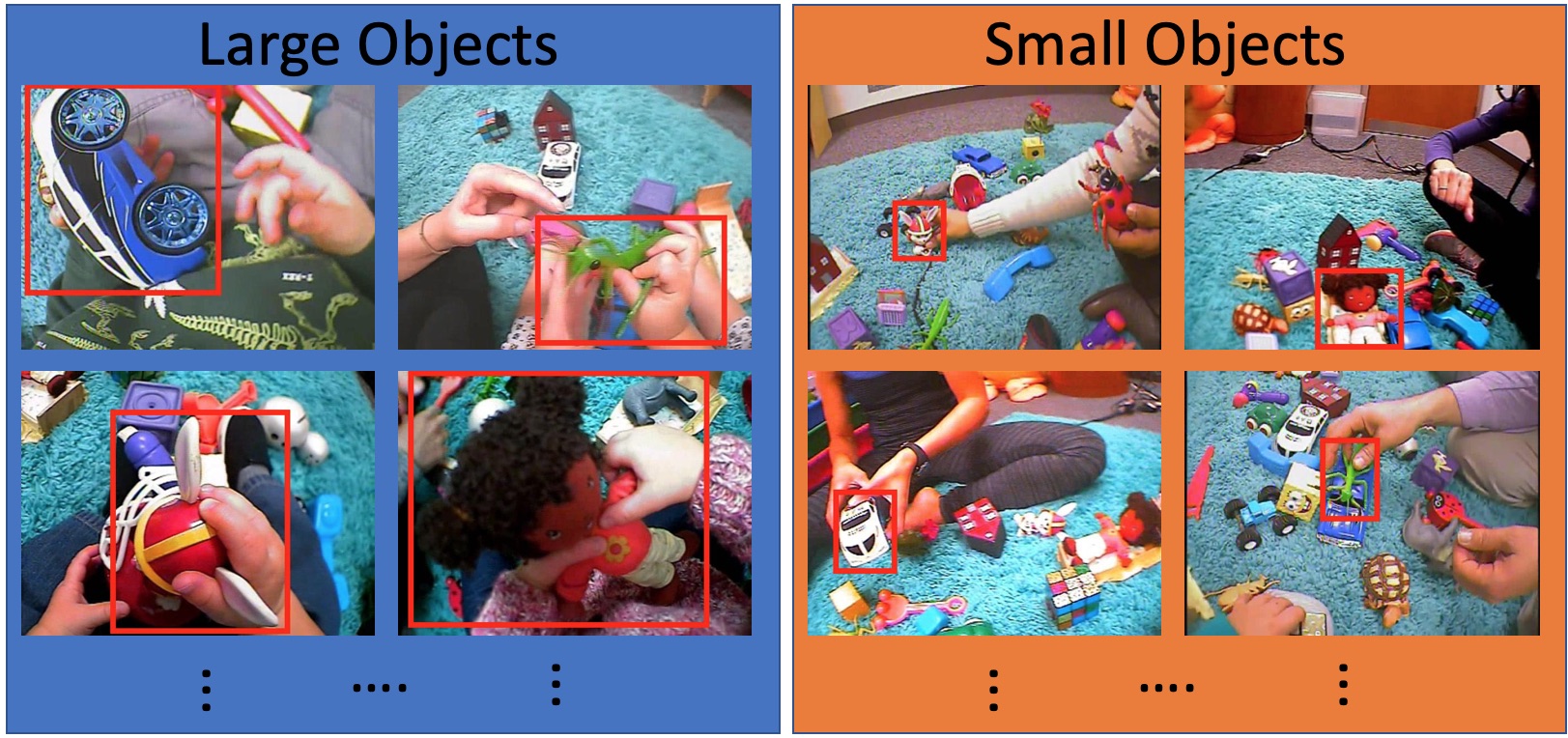}
	    \includegraphics[width=\linewidth]{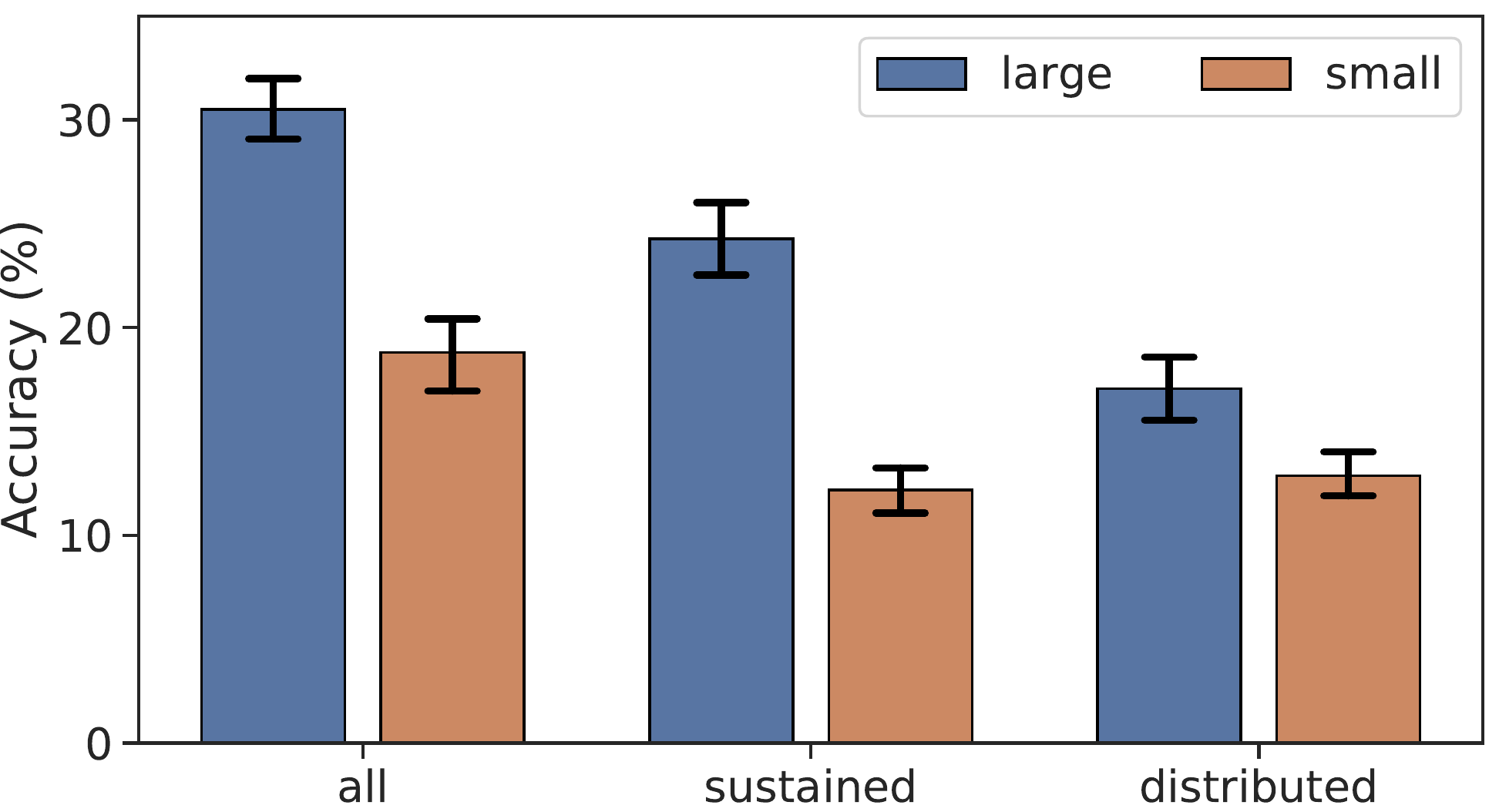}
        \caption{Effect of object size. Naming events
          were divided based on object size; instances in the
          large set contain visual instances of named objects large
          in view whereas the named objects are small in view in the
          small
          set. \label{fig:study4size} \label{fig:study4temp} \label{fig:sizesample}}
\end{figure}

If the target object in a naming event is large in view, that object
is more likely to be attended by infants. Thus, infants' sustained
attention on a target object is likely to co-vary with the size of the
object. If so, the difference in the learning results described
above could be due to sustained attention but not object size. To
distinguish the effects on word learning between those two co-varying
factors, we divided naming events into sustained attention and
distributed attention as in Study 2, and examined the effects of
object size in those two situations. In each case, we
used a median split to further divide naming events into a large
subset and a small subset. As shown in Figure~\ref{fig:sizesample},
when infants showed sustained attention on named objects, the model
trained based on large targets outperformed the same model trained
with small targets ($M_{large}=24.27\%, SE_{large}=2.03\%;
M_{small}=12.18\%, SE_{small}=1.5\%; \beta=0.37, t=4.,p<0.001$). In
the cases of naming events with distributed attention, the model again
favored events with large target objects over those
with small targets ($M_{large}=17.07\%,
SE_{large}=1.60\%; M_{small}=12.88\%, SE_{small}=1.33\%; \beta=0.20,
t=2.02,p<0.05$). Taken together, these results suggest that visual
properties of the target object during a naming event have direct and
unique influence on word learning.

\section{General Discussions}

Despite the fact that the referential uncertainty problem in word learning was
originally proposed as a philosophical puzzle, infant learners need to
solve this problem at the sensory level. From the infant's point of
view, learning object names begins with hearing an object label while
perceiving a visual scene having multiple objects in view.
However, many computational models on language learning use simple
data pre-selected and/or pre-cleaned to evaluate the theoretical ideas
of learning mechanisms instantiated by the models. We argue that to
obtain a complete understanding of learning mechanisms, we need to
examine not only the mechanisms themselves but also the data on which
those mechanisms operate. For infant learners, the data input to
their internal processes are those that make contact with their
sensory systems, so we capture the input data with egocentric video and head-mounted eye tracking. Moreover, compared to prior studies of word learning from third-person images~\cite{chrupala2015learning}, the present study is the first, to our knowledge, to use actual
visual data from the infant's point of view to reconstruct infants'
sensory experiences and to show how a computational model can solve the
referential uncertainty problem with the information available to
infant learners.

There are three main contributions of the present paper as the first
steps toward using authentic data to model infant word learning. First, our
findings show that the available information from the infant's point
of view is sufficient for a machine learning model to successfully
associate object names with visual objects. Second, our findings here
provide a sensory account of the role of sustained attention in early
word learning. Previous research showed that infant sustained
attention at naming moments during joint play is a strong predictor of
later vocabulary~\cite{yu2019infant}. The results here offer a
mechanistic explanation that the moments of sustained attention during
parent naming provide better visual input for early word learning
compared with the moments when infants show more distributed
attention.  Finally, our findings provide quantitative evidence on how
in-moment properties of infants' visual input influence early word
learning.

The present study used only naming utterances in parent speech (object
names in those utterances, etc.), but we know that parent speech during
parent-child interaction is more information-rich. For example, studies
show that individual utterances in parent speech are usually
inter-connected, forming episodes of coherent discourse
that facilitate child language
learning~\cite{suanda2016,frank2013social}. To better approximate
infants' learning experiences in our future work, we plan to include
both object naming utterances and other referential and
non-referential utterances as the speech input to computational
models. Including the whole speech transcription will also allow us to
examine how infants learn not only object names but also other types
of words in their early vocabularies, such as action verbs. In
addition, we know that social cues in parent-child interaction play a
critical role in shaping the input to infant learners. With egocentric
video and computational models, our future work will simulate and
analyze how young learners detect and use various kinds of social cues
from the infant's point of view. 

\section{Acknowledgment}
This work was supported in part by the National Institute of Child
Health and Human Development (R01HD074601 and R01HD093792), the
National Science Foundation (CAREER IIS-1253549), and the Indiana
University Office of the Vice Provost for Research, the College of
Arts and Sciences, and the Luddy School of Informatics, Computing, and
Engineering through the Emerging Areas of Research Project 
\emph{Learning: Brains, Machines and Children}.

\bibliographystyle{apacite}
\setlength{\bibleftmargin}{.125in}
\setlength{\bibindent}{-\bibleftmargin}

\bibliography{references}

\end{document}